\documentclass[a4paper]{article}

\usepackage[english]{babel}
\usepackage[utf8x]{inputenc}
\usepackage[T1]{fontenc}
\usepackage[round]{natbib}
\usepackage{csquotes}
\usepackage{wrapfig}
\usepackage{subcaption}

\usepackage[a4paper,top=3cm,bottom=2cm,left=3cm,right=3cm,marginparwidth=1.75cm]{geometry}

\usepackage{amsmath}
\usepackage{graphicx}
\usepackage[colorinlistoftodos]{todonotes}
\usepackage[colorlinks=true, allcolors=blue]{hyperref}

\newcommand{\ds}[1]{\textsc{#1}}

\title{A Critique of a Critique of Word Similarity Datasets:\\Sanity Check or Unnecessary Confusion?}
\author{Minh Le\\\texttt{m.n.le@vu.nl}}

\begin{document}
\maketitle

\begin{abstract}
Critical evaluation of word similarity datasets is very important for computational lexical semantics. This short report concerns the sanity check proposed in \cite{Batchkarov2016} to evaluate several popular datasets such as \ds{mc}, \ds{rg} and \ds{men} -- the first two reportedly failed. I argue that this test is unstable, offers no added insight, and needs major revision in order to fulfill its purported goal.
\end{abstract}

\section{Evaluating Evaluation Datasets}

Computational linguists rely on datasets to guide our exploration, judge our methods, and, practically speaking, get our papers published. But before we allow them to substitute our linguistic universe, it is only wise to give them a close examination.

It does not take a genius to know that small datasets are unreliable. 
Some popular datasets in lexical semantics certainly fall into this category: \cite{rubenstein.goodenough1965} and \cite{miller.charles1991} which contain only 65 and 30 examples respectively. It is well-known that any conclusions drawn from those datasets must be taken with a grain of salt but Batchkarov et al. make a stronger claim: 
\begin{displayquote}
``\ds{rg} and \ds{mc} do not sufficiently capture the degradation of vector quality as noise is added because $\rho$ may increase with $n$. The variance of the measurements is also very high. Both datasets therefore fail the sanity check.''
\end{displayquote}

\begin{wrapfigure}{R}{0.35\textwidth}
  \begin{center}
    \includegraphics[width=0.35\textwidth]{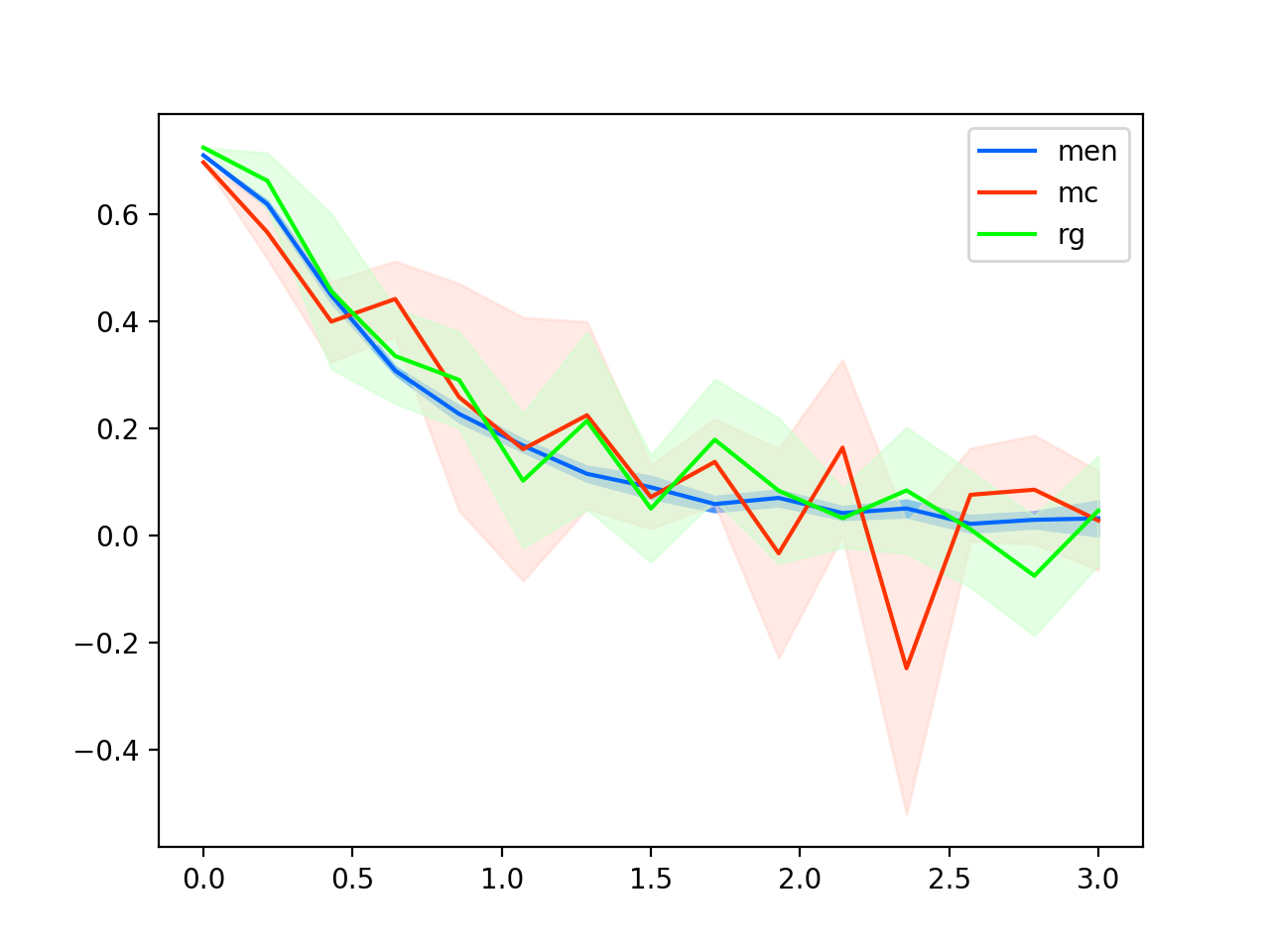}
  \caption{Correlation with similarity datasets vary with noise ($k=5$, horizontal axis: maximum random noise, vertical axis: Spearman's $\rho$).}\label{fig:repeats5}
  \end{center}
  \vspace{-4em}
\end{wrapfigure}

Here, the authors do not talk about uncertainty anymore but make a categorical decision. This can be likened to the $p$-value approach to hypothesis testing. In both cases, researchers force yes/no decision by checking some property: whether \textit{p} is less than a certain value or whether a plot is monotonic.

\section{Randomness and Instability}

One crucial difference between Batchkarov et al.'s sanity check and significance tests is that the computation of the former involves randomness while the later does not. To test sensitivity to noise, the authors sample random vectors and add them to the vectors predicted by \texttt{word2vec}, expecting that, as the magnitude of the random vectors increase, they will override the information inside \texttt{word2vec} vectors and drive Spearman's $\rho$ down. 

Figure \ref{fig:repeats5} demonstrates this intuition using synthetic data. I use synthetic datasets of the same size as those reported in Batchkarov et al.\footnote{Here, \textit{men} is a synthetic dataset of the same size as \ds{men} \citep{Bruni2014}, etc. Notice that we obtain a graph very similar to Figure~3 in Batchkarov et al.} and follow their procedure by adding noise of increasing magnitude to synthetic word vectors. The number of samples $k$ (i.e.\ the times noise vectors are sampled and performance is measured) is not reported in their paper so I start with a small number $k=5$. It is immediately clear from the figure that similarity scores tend to go down as error increases, though with occasional surges.

Batchkarov et al. go further, and I think it is here that they get it wrong, by claiming that correlation should always go down:
\begin{displayquote}
``Model performance, as measured by the evaluation method, should be a monotonically decreasing function of the amount of noise added.''
\end{displayquote}
In other words, they consider the surges of scores as not an artifact of the randomness in their methodology but an unwanted property of the dataset and those that demonstrate such surges are considered to ``fail'' the test.

\begin{figure}
    \centering
    \begin{subfigure}[b]{0.3\textwidth}
        \includegraphics[width=\textwidth]{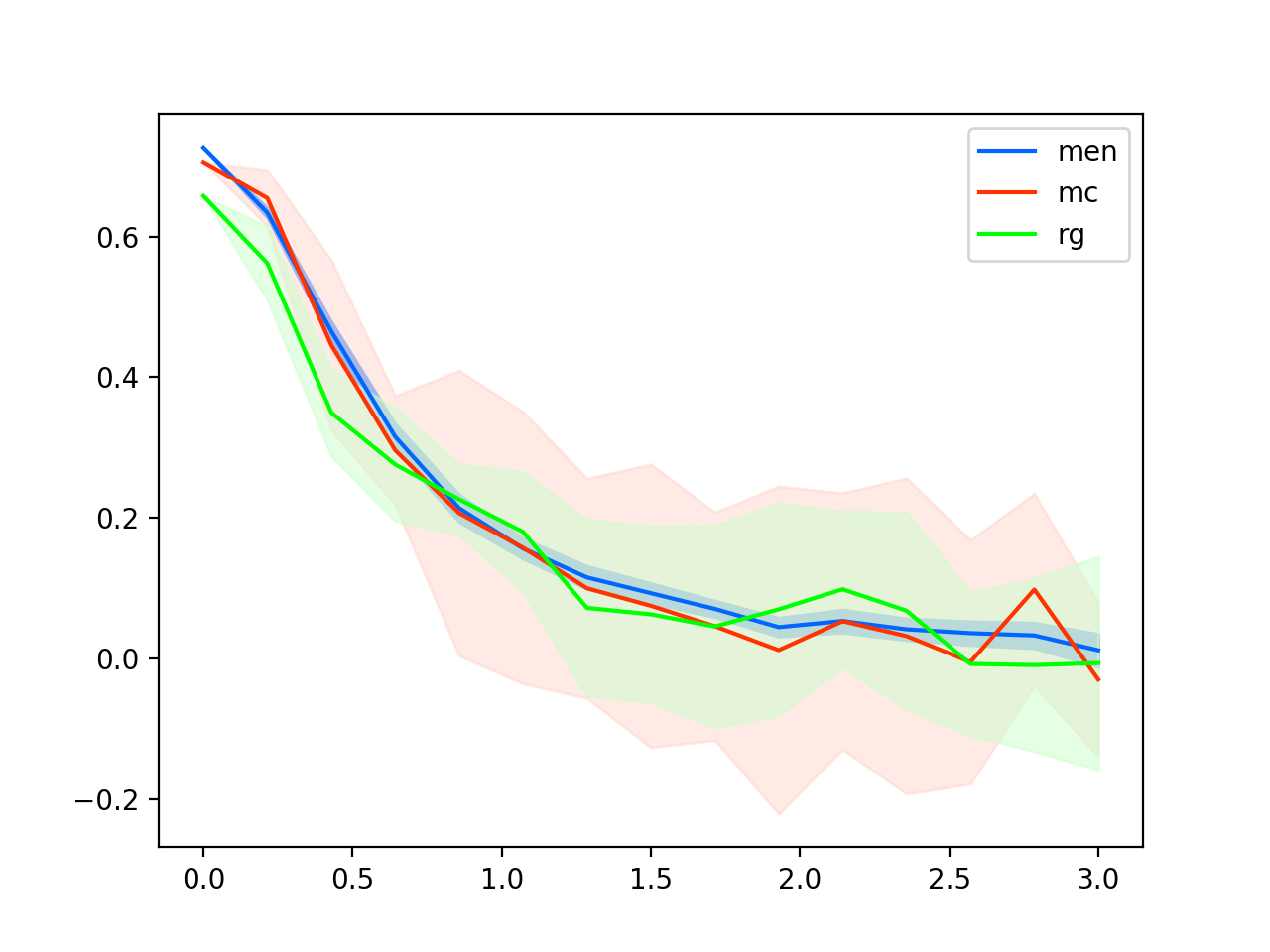}
        \caption{$k=10$}
        \label{fig:repeats10}
    \end{subfigure}
    ~ 
    \begin{subfigure}[b]{0.3\textwidth}
        \includegraphics[width=\textwidth]{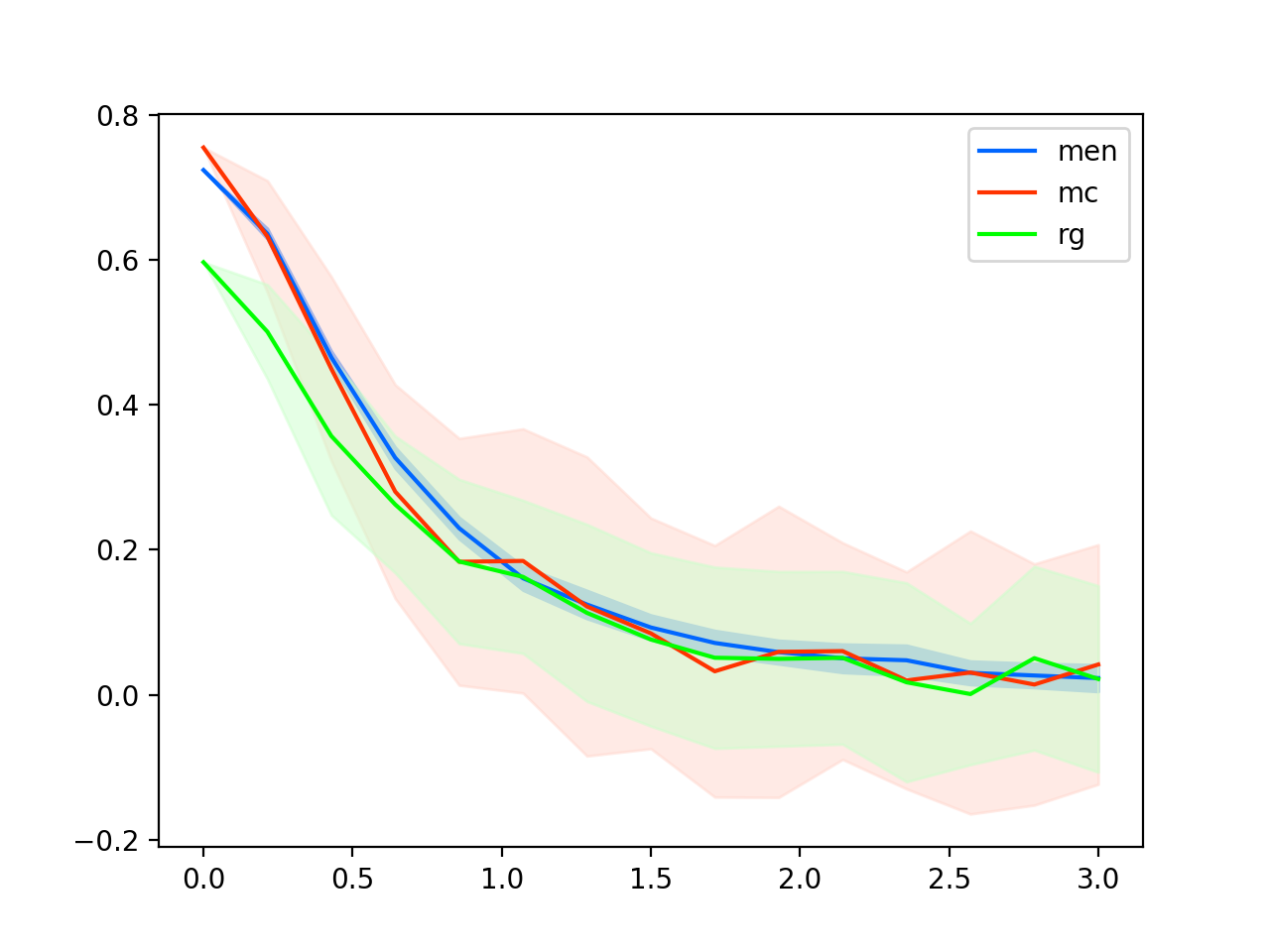}
        \caption{$k=50$}
        \label{fig:repeats50}
    \end{subfigure}
    ~ 
    \begin{subfigure}[b]{0.3\textwidth}
        \includegraphics[width=\textwidth]{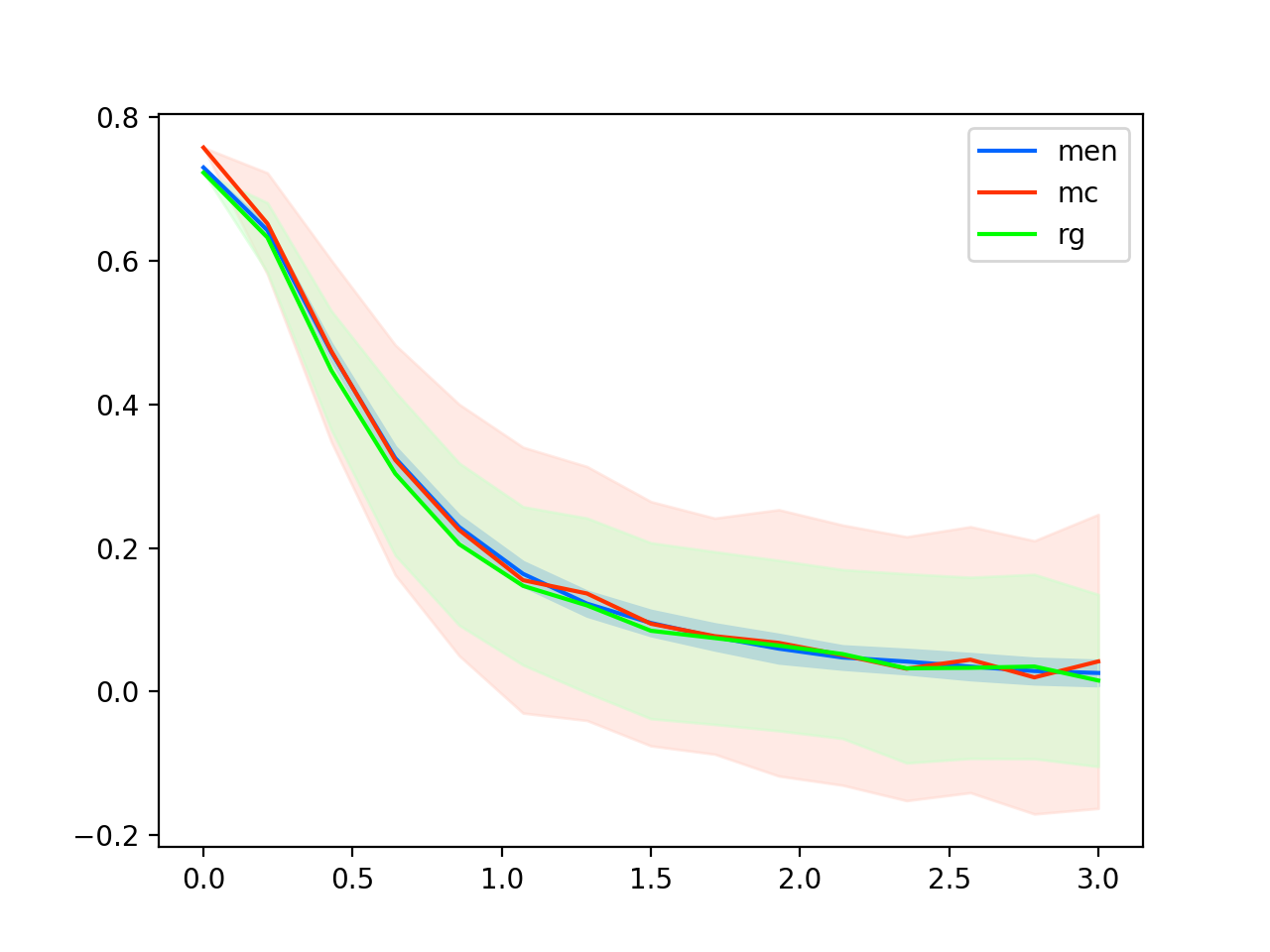}
        \caption{$k=250$}
        \label{fig:repeats250}
    \end{subfigure}
    
    \begin{subfigure}[b]{0.3\textwidth}
        \includegraphics[width=\textwidth]{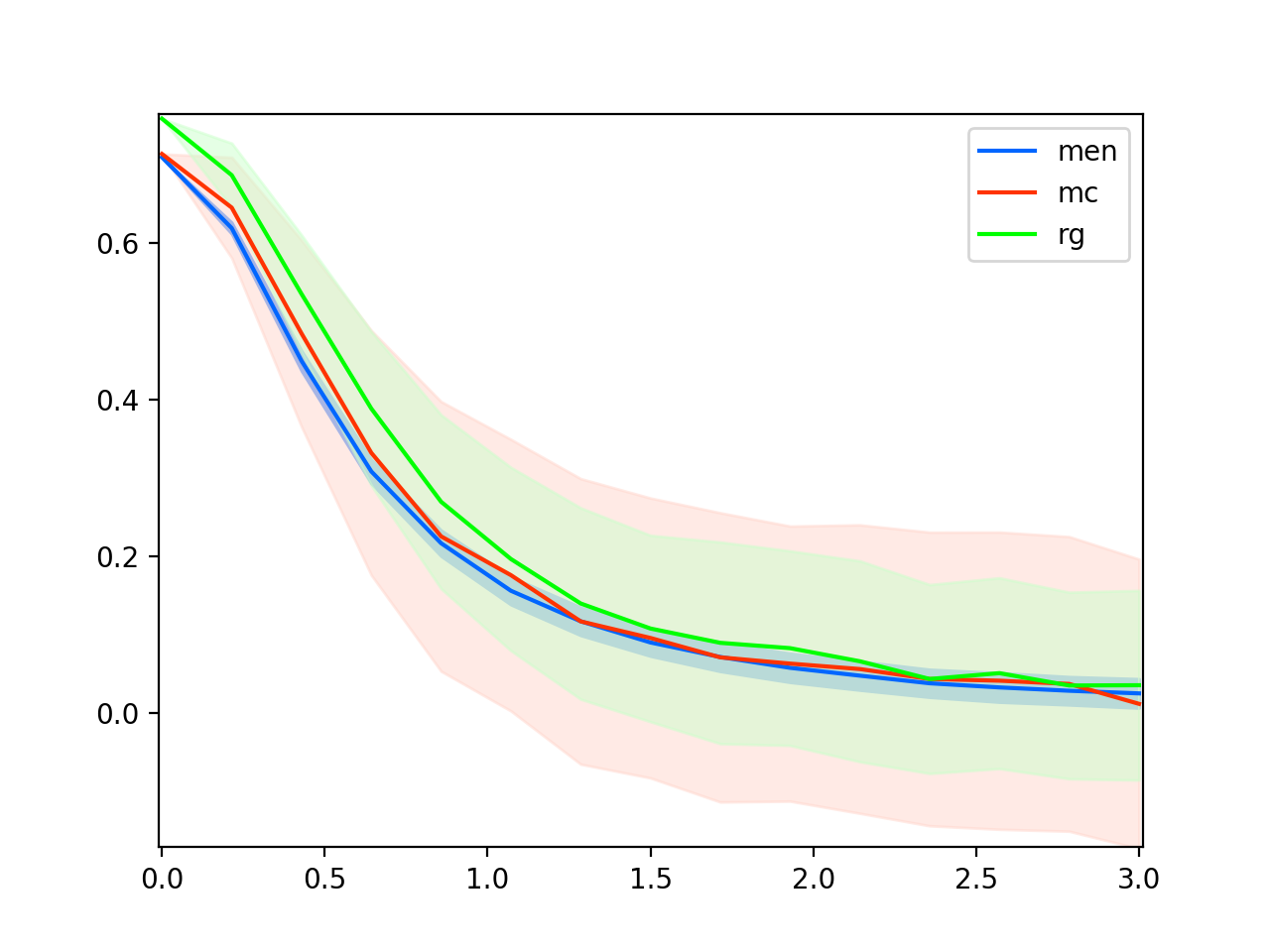}
        \caption{$k=500$}
        \label{fig:repeats500-mc-ok}
    \end{subfigure}
    \begin{subfigure}[b]{0.3\textwidth}
        \includegraphics[width=\textwidth]{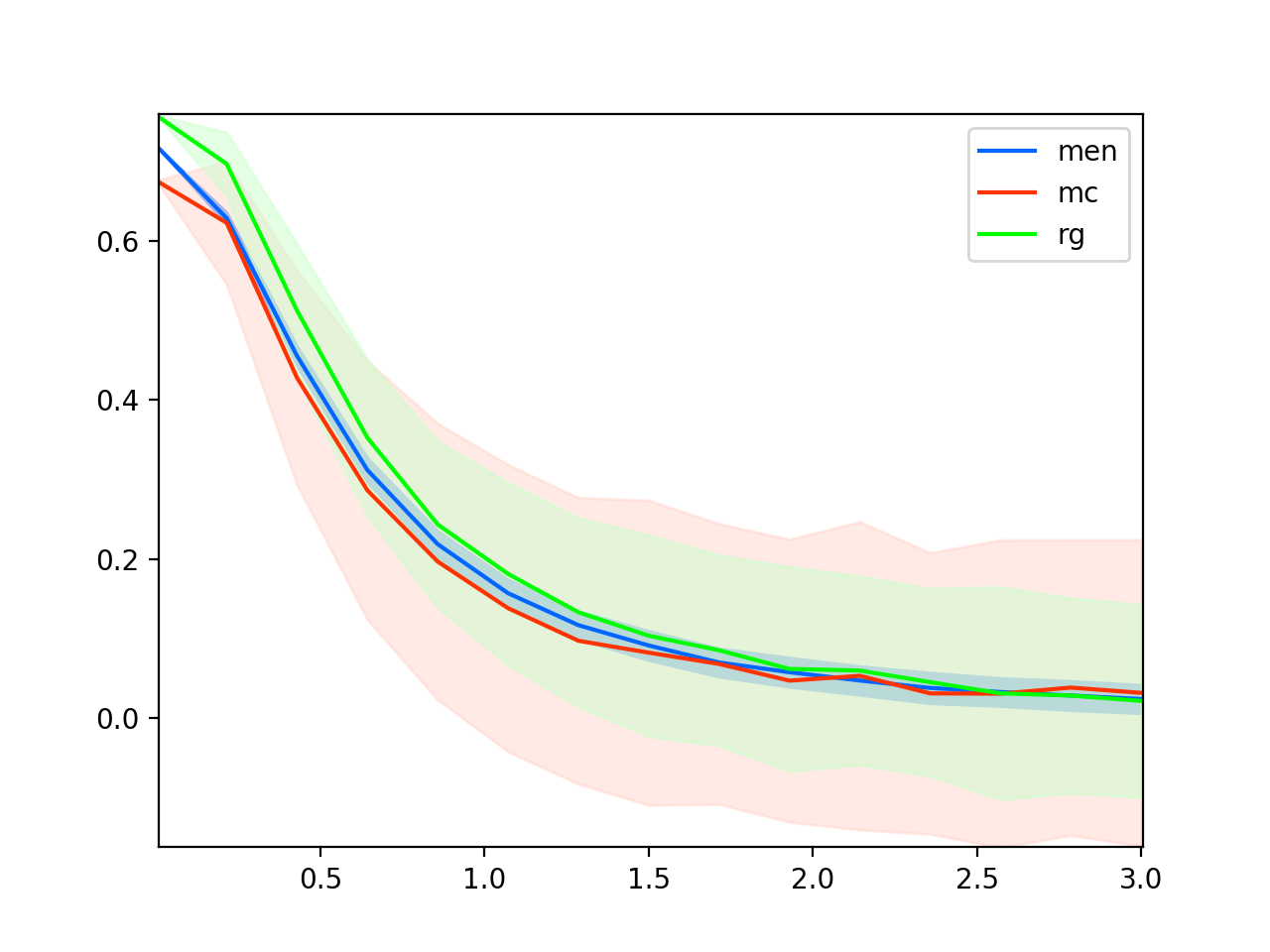}
        \caption{$k=500$}
        \label{fig:repeats500-rg-ok}
    \end{subfigure}
    \caption{Varying shape of the plot depending on the number of samples.}\label{fig:samples}
\end{figure}

It is not hard to show the contrary though. By increasing $k$ from 5 to 500 (Figure~\ref{fig:samples}), one can observe that the lines of \ds{mc} and \ds{rg} smooth out and get closer and closer to that of \ds{men}. At $k=500$, depending on the particular run, we can get a perfectly monotonically decreasing line for \ds{mc} (Figure~\ref{fig:repeats500-mc-ok}) or \ds{rg} (Figure~\ref{fig:repeats500-rg-ok}). In other words, sometimes \ds{mc} fails the sanity check and sometimes it passes; the same goes for \ds{rg} -- the test is instable.

\section{Conclusions}

\cite{Batchkarov2016} proposed a sanity check of word similarity datasets, among other things. In this report, I argue and demonstrate by synthetic data that their test leads to inconsistent conclusions and therefore should not be used. The root cause appears to be confusion between tendency and monotonicity and wrong attribution of abnormality. Moreover, the behavior of the test appears to reflect solely the size of the datasets it is run on. To go beyond an illustration of the obviously high variance of measurement on small datasets and become a viable quality assurance tool, the test needs to be revised to respond to factors such as the word frequency, similarity range, the relationship between included words, etc.

The source code of the experiments is available at: \url{https://git.io/vQMAS}.

\bibliographystyle{plainnat}
\bibliography{mybib}

\end{document}